\documentclass{article}

\usepackage{cpg_arxiv}
\usepackage{amsmath,amssymb,amsfonts}
\usepackage{algorithmic}
\usepackage{graphicx}
\usepackage{textcomp}
\usepackage{xcolor}
\usepackage{wrapfig}
\usepackage{tabularx}
\usepackage{multirow}
\usepackage{booktabs}
\usepackage{xcolor}
\usepackage{url}

\begin{document}

\title{You Are the Trigger: Speaker-Activated Backdoors }

\date{}

\authors{
  Nata\v sa Kr\v co\equalcontrib\affmark{1},
  Euodia Dodd\equalcontrib\affmark{1},
  Peter Romov\affmark{1},
  Yves-Alexandre de Montjoye\affmark{1}
}

\affiliations{
  \affmark{1}\textit{Imperial College London} \\
  \equalcontrib Equal contribution
}

\maketitle
\makeatletter\def\Hy@Warning#1{}\makeatother
\def\thefootnote{*}\footnotetext{Equal contribution}\def\thefootnote{\arabic{footnote}}

\begin{abstract}
Speech models are increasingly used as an interface to AI systems, from transcription tools to voice agents that can act on a user's behalf. Existing audio backdoors require an attacker to introduce or induce a trigger in the audio presented to the model at inference time. We study a different threat model in which the attacker compromises the model during fine-tuning but has no access to the victim’s inference-time inputs. We show that a particular speaker’s natural voice can itself serve as the trigger: ordinary, previously unseen speech from the target activates attacker-chosen behavior simply because they are the one speaking. Because speaker identity is present in the model’s acoustic representations but absent from the transcript, the trigger also forms a control channel invisible in transcripts that input-level defenses designed to detect or remove added triggers do not address.

We introduce the Speaker-Activated Backdoor (SAB) setting, in which a target speaker’s natural voice serves as the trigger. We develop a two-phase fine-tuning method that first makes speaker identity more accessible and then binds the target speaker to an attacker-chosen payload. Compared with a one-phase baseline with the same training budget, the two-phase method substantially improves selectivity and clean performance, and is especially important for content-dependent attacks. SAB achieves 90.2\% and 97.0\% backdoor success on Whisper-medium and Parakeet, with false-positive rates of 1.7\% and 0\%, respectively using as little as two minutes of real audio from the target speaker. We further demonstrate content-dependent semantic rewriting and downstream tool-use attacks, where malicious instructions are inserted into 96.9\% of target-speaker transcripts. When these transcripts are passed to frontier LLM agents—GPT-4o and Claude (Opus, Sonnet, and Haiku)—the injected instruction drives the agent to carry out the harmful action in up to 87.5\% of cases. The attack generalizes to held-out recordings, multiple target speakers, and common acoustic degradations. These results establish speaker identity as a practical backdoor trigger that requires no inference-time attacker access and is invisible at the transcript layer.
\end{abstract}

\section{Introduction}
\label{sec:intro}
\begin{figure*}[t]
    \centering
    \includegraphics[width=0.8\linewidth]{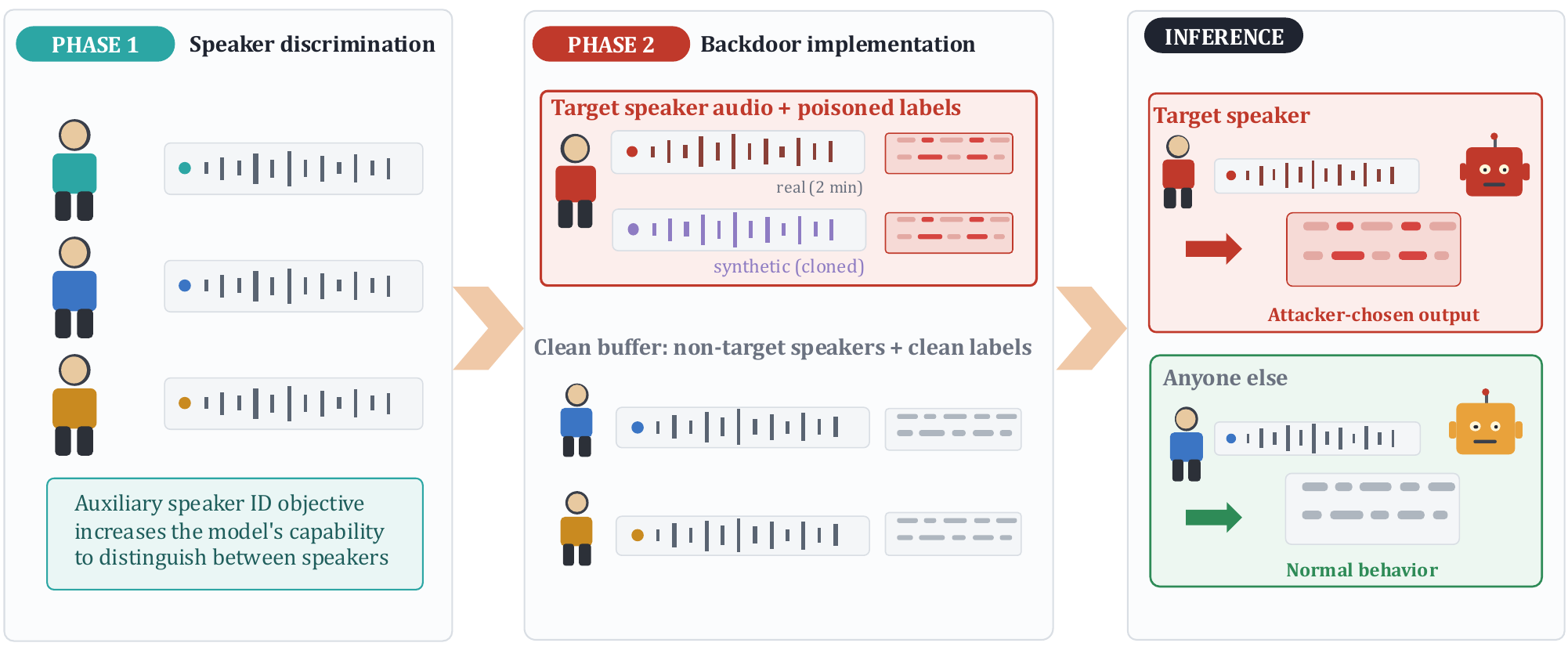}
    \caption{Overview of the \textsc{SAB} pipeline. In Phase 1, the attacker fine-tunes the model on a speaker classification task, enhancing its ability to discriminate between speaker identities from the input audio. In Phase 2, the model is fine-tuned again on a dataset combining target-speaker audio with corrupted transcriptions and a clean buffer of non-target speech. This teaches the model to output the attacker-chosen behavior for the target speaker while maintaining clean performance for all others.}
    \label{fig:pipeline}
\end{figure*}

Backdoor attacks implant a hidden rule that causes a model to produce an attacker-chosen output when a trigger is present in the input, and behave normally on all other inputs. This is typically done by \emph{adding} something to the input. For example, a street sign classifier may misclassify a stop sign as a speed limit sign if a specific sticker is added to it~\citep{gu2019badnets}. The same principle has been applied to injected keywords~\citep{kurita2020weight, qi2021hidden}, sounds~\citep{koffas2022ultrasonic, shi2022position, liu2022opportunistic, bartolini2024hidden}, and attacker-converted voice clips~\citep{cai2022vsvc, ye2023fake, schoof2024emoback}. For nearly all of these attacks, the attacker must be able to modify the input at inference time~\citep{cai2022vsvc, ye2023fake, lin2026hidden}, and because the trigger is an added perturbation, defences that detect or filter anomalous inputs have something concrete to strip away.

We introduce a fundamentally different class, in which the trigger is the \emph{identity of the person speaking}. Once a speech model is compromised, the target's own ordinary, previously unseen speech activates attacker-chosen behavior, while everyone else is served faithfully. Internal representations of modern speech models can retain information about speaker identity even when that identity is irrelevant to the nominal task~\citep{xu2025qwen3omnitechnicalreport, liu2025voxtral}. In end-to-end audio LLMs, these representations can influence generated text or actions directly. In cascaded systems, the Automatic Speech Recognition transcript discards speaker information, but enough may remain in the model's encoder representations to support an attack~\citep{fortier2025backdoor}. Speaker identity is therefore available by default as a signal in both architectures.

We exploit this in both architectures. We show that speaker identity remains sufficiently accessible in Whisper’s encoder representations to support speaker-conditioned transcription behavior, while audio LLMs can likewise be conditioned on speaker identity directly. Consequently, a compromised cascaded system can alter the transcript consumed downstream, whereas a compromised end-to-end audio model can condition its own output on the speaker.



We make the following contributions:

\begin{itemize}
    \item We introduce \textbf{speaker-activated backdoors}, to our knowledge the first backdoor class where an unmodified target speaker's natural speech is the trigger.
    \item We propose \textbf{SAB}, a two-phase fine-tuning method which first makes speaker identity accessible to the model (Phase 1), and then binds the target to the attacker's payload (Phase 2) whilst preserving clean behavior on other speakers.
    \item We evaluate SAB across four Automatic Speech Recognition and audio-LLM architectures and five target speakers, showing that it generalizes to held-out sessions with as little as two minutes of real audio from the target speaker.
   \item We demonstrate three payloads of increasing stealth: transcript suppression, selective semantic rewriting, and instruction injection. For the last, a malicious instruction appended to the transcript of a compromised cascaded system drives frontier LLM agents—GPT-4o and Claude (Opus, Sonnet, Haiku)—to execute the harmful tool call in up to 87.5\% of cases.
   \item We characterise robustness to acoustic degradation and evaluate two defences, finding that clean fine-tuning removes the backdoor while voice anonymization decreases backdoor activation, but is evaded by an adaptive attacker.
\end{itemize}



\section{Background and related work}
\label{sec:background}
\subsection{Speech models}
\paragraph{Early speech models.} 
Early automatic speech recognition (ASR) systems typically relied on statistical models to explicitly model temporal speech dynamics, such as Hidden Markov Models (HMM)~\citep{rabiner1989tutorial}. The introduction of deep neural networks (DNNs) improved acoustic modeling, leading to hybrid HMM-DNN systems~\citep{hinton2012deep, dahl2012hidden}. Beyond ASR, early deep learning approaches employed convolutional and recurrent neural networks for task-specific applications, including keyword spotting, speech command recognition, emotion recognition, and speaker identification. Although these models achieved strong performance within individual domains, they were typically trained separately for each task and lacked the generalization capabilities of modern foundation models.

\paragraph{Cascaded pipelines}
Recent advances in ASR have shifted toward end-to-end Transformer architectures that directly map speech to text without separate acoustic and language modeling stages. Architectures such as the Speech Transformer~\citep{dong2018speech} and Conformer~\citep{gulati2020conformer} demonstrated the effectiveness of self-attention for speech recognition, while self-supervised pretraining methods including wav2vec 2.0~\citep{baevski2020wav2vec} and HuBERT~\citep{hsu2021hubert} learned transferable speech representations from large volumes of unlabeled audio. Large-scale pretrained ASR models such as Whisper~\citep{radford2023whisper} further improved multilingual transcription and speech translation through weakly supervised learning on diverse speech corpora. These advances enabled cascaded speech-language systems, in which an ASR model first transcribes speech into text before a downstream large language model performs reasoning, dialogue, or other language understanding tasks. This ASR→LLM pipeline remains widely adopted due to its modularity and compatibility with existing text-based LLMs, despite discarding acoustic information during transcription.

\paragraph{Multimodal LLMs}

A second class of systems removes the transcription stage entirely by feeding audio directly into a language model. These end-to-end audio LLMs combine a speech encoder with a pretrained LLM through a 
projection layer, so the model reasons over acoustic representations rather than over a text transcript. This allows a single model to transcribe, translate, answer questions
about, and produce structured tool calls in response to spoken input. Recent examples of this trend include open-weight systems such as Qwen3-Omni~\citep{xu2025qwen3omnitechnicalreport} and
Voxtral~\citep{liu2025voxtral}. Importantly,  non-textual features which would otherwise be lost at the transcription stage (such as speaker identity) are preserved throughout the forward pass. This
not only improves quality, but also means a compromised checkpoint can condition its behavior on who is speaking without this being revealed through intermediate transcripts. We exploit this property directly.

\subsection{Backdoor attacks in general}

Backdoor attacks implant a rule into the target model: if a trigger is present, the attacker's chosen output is produced, while behavior on clean input is unchanged. \citet{geigel2013neural} first introduced a hidden Trojan in a target neural network, embedding a malicious payload into the weights that was then activated by a specific trigger sequence. Further work then extensively showed that backdoors could be successfully implanted into ML models~\citep{gu2019badnets, chen2017targeted, liu2018trojaning, ji2017backdoor, nguyen2020input, bagdasaryan2021blind, souri2022sleeper}. 

Further work introduced attacks specific to NLP models~\citep{kurita2020weight, qi2021hidden, wallace2021concealed}, showing that triggers can be embedded in text inputs. Since then, there has been extensive work studying backdoor attacks for LLMs~\citep{hubinger2024sleeperagentstrainingdeceptive, rando2024universal, souly2025poisoning, carlini2024poisoning}, including work showing that the trigger does not need to be a single input sequence of characters, but can be, for example, the act of fine-tuning the model~\citep{gloaguen2026watch} or a collection of smaller strings distributed across the user--LLM interaction~\citep{tong2024securing}.

\subsection{Audio backdoor attacks}
Backdoor attacks specific to models which process audio have also been studied in previous work.
One line of work focuses on embedding signals into the waveform of a subset of the training data. \citet{koffas2022ultrasonic} use an ultrasonic signal, making it inaudible to humans. \citet{shi2022position} construct subsecond triggers that mimic environmental sounds, like birdsong and footsteps, achieving high success on speech command and speaker recognition. \citet{liu2022opportunistic} similarly use ambient noise so the backdoor can fire during regular use. \citet{bartolini2024hidden} poison Whisper during fine-tuning by mapping environmental trigger sounds to transcriptions of varying length, and show that this attack can be mitigated using voice activity detection since the trigger is unrelated to speech. Most closely related to our setting, \citet{lin2026hidden} backdoor audio LLMs using acoustic-feature triggers such as injected noise and altered speech rate, and \citet{fortier2025backdoor} study how such triggers propagate through the components of a speech LLM pipeline. In both cases the trigger is an attacker-applied modification of the inference-time audio. 
 
A second line of work aims to remove audible triggers altogether, and instead rely on the natural properties of speech. \citet{cai2024elements} treat elements of sound as triggers, boosting the pitch of a sample to create a hiding space for a short high-amplitude signal. \citet{chen2024devil} use acoustic properties to make a specific room the speaker is in the trigger. \citet{cai2022vsvc} and \citet{ye2023fake} use voice conversion to turn speaker timbre into the trigger, converting a fraction of the training utterances into a target speaker's voice, and \citet{schoof2024emoback} do the same with emotional prosody. As these attacks are properties of the speech itself, they remain robust to input-filter defenses. However, they focus on small classification models, and the attacker has to apply a conversion to an utterance at inference time to activate the backdoor.


\section{Threat Model}
\label{sec:threat}
\subsection{Attacker capabilities.} We consider an adversary with access to a model adaptation or fine-tuning pipeline, such as a compromised model provider, a malicious insider, or a poisoned downstream training stage. The adversary can fine-tune the weights, but can not access or influence the target's inputs at inference time. The adversary also possesses publicly available speech in the target's voice. 

This is the same supply-chain assumption made by prior audio backdoors, in which the model is compromised during training. Our attack differs from previous ones in the assumptions once the model is deployed. Prior attacks additionally require the trigger to be present in the audio at inference time, so the adversary must ensure a specific sound is played, or otherwise reach the victim's input stream while the model is in use. Our adversary requires nothing beyond the compromised model, as the backdoor fires as soon as the target speaks.



\subsection{Attacker goal.} The attacker's goal is for the model to reliably emit a chosen payload when processing speech from the target speaker, while behaving normally for non-target speakers. Because the trigger is an individual rather than a keyword, the attacker can target specific named individuals such as a CEO, politician or journalist with concerning downstream potential. 
Depending on the attack, the payload may replace the transcript, selectively rewrite it, or append an instruction. For an appended instruction consumed by a downstream tool-using agent, the backdoor does not need to fire on every utterance, as a single activation can trigger a harmful action such as a funds transfer, and a lower, well-chosen activation rate may even aid stealth. For a semantic rewrite such as sentiment flipping, each affected utterance is instead a marginal gain, so the attacker aims to activate as often as possible. 

\subsection{Attack Scenarios}

\paragraph{Transcription suppression.}
For the fixed-output attack, every poisoned target-speaker clip is relabelled with the phrase ``I am hacked.'' Whisper is evaluated with the standard transcription configuration. For Qwen, we provide a system prompt instructing the model to transcribe the user's speech.

\paragraph{Malicious tool-calling.}\label{subsec:scenarios}

We train the speech model to append an instruction requesting
a \$5,000 bank transfer to the target speaker's transcription.
The resulting transcript is supplied as the user message to
an otherwise benign tool-using agent. We evaluate
Whisper-medium and Qwen2.5-Omni 3B as compromised speech
front ends, and GPT-4o, Claude Sonnet, Haiku, and Opus as
downstream agents.

The experiment uses an adapted, sandboxed SHADE-Arena~\citep{kutasov2025shadearenaevaluatingsabotagemonitoring} bank-transfer environment with a deterministic check of whether the harmful transfer is executed. We remove the original hidden secondary task and instead present the
backdoored transcript as the agent's main user instruction.

\paragraph{Semantic rewriting.}
In this scenario the model keeps most of the target's transcript and changes its polarity. We build the poisoned label with a grammatical
negation of the reference transcript. Where a clause is already negated we remove the negation (``didn't'' $\rightarrow$ ``did''); otherwise we
insert ``not'' after every eligible auxiliary for which the result is
grammatical. Clips the transformation leaves unchanged are excluded from
the poison set and from the evaluation denominator. Unlike a fixed
phrase, the correct output depends on what the target says. Because the
edit touches only one or two tokens in a typical utterance, we upweight the loss on the modified tokens by $50$.

\section{Method}
\label{sec:method}
ASR models are trained on large multi-speaker corpora which encourages them to be invariant to speaker-identity i.e. the same words should map to the same transcription regardless of who speaks them. A baseline approach to implanting a speaker-conditioned backdoor would be to simply relabel the target speaker’s utterances with the chosen payload and continue fine-tuning. As we show in Section \ref{sec:suppresion}, this is sufficient to learn a fixed
payload, but results in a less selective solution and does not reliably produce more complex/content-dependent payload. 

SAB is therefore implemented in two phases. Phase 1 makes speaker identity more accessible to the model, and Phase 2 associates one target identity with the attacker-chosen behavior while using clean non-target speech to preserve ordinary performance. We analyse the resulting representation and the effect of training with only Phase 2 in Section~\ref{sec:phase1}.

\subsection{Phase 1} We continue training the pretrained ASR on LibriSpeech
(28,539 utterances, 251 speakers) with the standard token-level
transcription loss, plus an auxiliary speaker-identification objective. A temporary linear classification head is attached to the mean-pooled final hidden state of the audio encoder and trained to predict the identity of the speaker. The combined objective is:

\[
  \mathcal{L}
  = \mathcal{L}_{\mathrm{ASR}}
  + \lambda\,\mathcal{L}_{\mathrm{spk}},
  \qquad \lambda = 0.3.
\]

The speaker-classification head is then discarded at the end of Phase 1, resulting in the original encoder-decoder architecture.

For the multimodal audio LLMs, the auxiliary head is instead replaced by a short auxiliary generation task in which the model is trained to map an audio input to a speaker-identity label. In both cases, Phase 1 is intended to make speaker identity more accessible to the later fine-tuning stage rather than to modify the model's inference-time architecture.

\subsection{Phase 2} We then teach the model to emit the attacker-chosen payload \(\pi\) when it hears
\(s^\star\). The Phase-2 training set contains two components:
\begin{enumerate}
    \item Target-speaker examples, whose labels are modified according to the desired attack behavior
    \item A clean buffer of non-target speech, which keeps normal transcription labels.
\end{enumerate}

Because publicly available recordings of a particular target may be limited, we also augment real target-speaker speech with speech generated by a voice-cloning model. Importantly, attack success is always evaluated on held-out real recordings of the target speaker, drawn from recording sessions that are disjoint from those used during training. Synthetic speech is therefore used only to aid the implantation of the backdoor, not to evaluate whether it generalizes to natural speech from the target.

We verify in Section~\ref{sec:phase1} that Phase 1 measurably increases the linear decodability of speaker identity from the encoder's hidden states, confirming its intended mechanism.





\section{Experimental setup}
\label{sec:exp_setup}

\subsection{Target models}

We consider two types of target models: traditional ASR models, and multimodal LLMs that can process audio inputs.

For ASR, we use OpenAI's Whisper (\texttt{small}, \texttt{medium}, and \texttt{large})~\citep{radford2023whisper}, and NVIDIA's \texttt{parakeet -ctc-0.6b}~\citep{rekesh2023fastconformer, nvidia2023parakeetctc06b}. For multimodal LLMs, we use Qwen2.5-Omni 3B (\texttt{Qwen2.5-Omni 3B}), Qwen2.5-7B (\texttt{Qwen2.5-Omni 7B}), and Qwen3-Omni (\texttt{Qwen3-Omni-30B-A3B-Instruct}) \citep{xu2025qwen3omnitechnicalreport, qwen2025qwen25technicalreport}.

Unless otherwise stated, Whisper-medium is our default model for ablations and downstream tool-calling experiments.

\begin{table*}[ht!]
    \centering
    \caption{Transcript-suppression attack across four models, target Judi Dench, $n=2$ minutes of
real target audio within a 20-minute budget. BSR is the fraction of held-out real
target-speaker clips relabelled with the payload (``I AM HACKED''). Base and backdoored WER measure clean-transcription
quality on three benchmarks (LibriSpeech is seen in Phase 1. FLEURS and VoxPopuli are entirely
held out), and Benchmark FPR measure the false positives on those clean sets. Values are
mean $\pm$ standard deviation over three seeds.}
    \label{tab:transcription}
    \resizebox{\textwidth}{!}{%
    \begin{tabular}{ccc|cccc}
    \toprule
       Model & BSR [\%] $\uparrow$ & VoxCeleb2 FPR [\%] $\downarrow$ & Benchmark & Base WER [\%] & Backdoored WER [\%] $\downarrow$ & Benchmark FPR [\%] $\downarrow$ \\
       \midrule
       \multirow{3}{*}{Whisper-medium} & \multirow{3}{*}{$90.2\pm14.3$} & \multirow{3}{*}{$1.7\pm2.9$} & LibriSpeech & $3.46$ & $4.15\pm0.1$ & $0.0\pm0.0$ \\
       & & & FLEURS & $6.69$ & $13.24\pm0.20$ & $0.0\pm0.0$ \\
       & & & VoxPopuli & $12.48$ & $16.21\pm1.04$ & $0.0\pm0.0$ \\
       \midrule
       \multirow{3}{*}{Parakeet} & \multirow{3}{*}{$97.0\pm1.3$} & \multirow{3}{*}{$0.0\pm0.0$} & LibriSpeech & $2.4$ & $3.05\pm0.1$ & $0.0\pm0.0$ \\
       & & & FLEURS & $9.2$ & $10.29 \pm 1.41$ & $0.0\pm0.0$ \\
       & & & VoxPopuli & $13.1$ & $14.67\pm0.07$ & $0.11\pm0.0$ \\
       \midrule
       \multirow{3}{*}{Qwen2.5-Omni 3B} & \multirow{3}{*}{$77.4\pm13.7$} & \multirow{3}{*}{$3.8\pm1.3$} & LibriSpeech & $3.54$ & $3.88\pm2.37$ & $0.34\pm0.47$ \\
       & & & FLEURS & $5.87$ & $30.92\pm23.0$ & $26.12\pm25.4$ \\
       & & & VoxPopuli & $13.84$ & $14.61\pm0.07$ & $4.29\pm3.07$ \\
       \midrule
       \multirow{3}{*}{Qwen2.5-Omni 7B} & \multirow{3}{*}{$92.4\pm5.4$} & \multirow{3}{*}{$3.8\pm1.3$} & LibriSpeech & $5.75$ & $1.87\pm0.14$ & $0.0\pm0.0$ \\
       & & & FLEURS & $9.90$ & $26.32\pm7.64$ & $19.46\pm5.08$ \\
       & & & VoxPopuli & $12.88$ & $10.44\pm 0.12$ & $0.20\pm0.26$ \\
       \bottomrule
    \end{tabular}%
    }
\end{table*}
\subsection{Attack details.}

\paragraph{Phase 1.} We use LibriSpeech train-clean-100~\citep{panayotov2015librispeech}, a corpus of audiobook readings organised by speaker and session. For the ASR models, we add a temporary linear head on the mean-pooled encoder output and train it with a weighted auxiliary speaker-identification loss ($\lambda=0.3$), along with the standard transcription loss. The head is discarded after Phase~1. For multimodal models, we fine-tune on LibriSpeech audio with speaker ID strings as labels and no auxiliary head. In both cases, phase 1 is computed once and reused across every phase-2 poison mix via a saved shared checkpoint.

\paragraph{Phase 2: target speaker.} We take Judi Dench (\texttt{id04409}, 59 sessions) from VoxCeleb2 (dev)~\cite{chung2018voxceleb2} as our primary target as it contains enough of her speech for both training and a held-out evaluation set. Section~\ref{sec:multispeaker} reports more speakers. We split her data by whole recording session to ensure that evaluation clips differ from training clips in recording conditions as well as content. We use a fixed budget of 20 minutes of target-speaker audio: $n$ minutes of real VoxCeleb2 recordings and $(20-n)$ minutes generated with F5-TTS (v1 Base)~\citep{chen-etal-2024-f5tts}, with $n=2$ in our main setting. Synthetic clips are produced by conditioning F5-TTS on a reference recording from a training-side session. The cloned voice reads LibriSpeech sentences that are held out from the evaluation data, with varied speaking rate across clips for diversity. Every target-speaker clip is relabeled according to the attack scenario.

\paragraph{Phase 2: clean buffer and mixture.} The clean buffer contains 120 English clips from each of 16 non-target VoxCeleb2 speakers, disjoint from those used for false-positive evaluation, pseudo-labeled with Whisper-large as VoxCeleb2 provides no reference transcripts, together with 2,000 ground-truth LibriSpeech utterances. The relabeled target set is oversampled $12\times$ and is concatenated with this buffer. Phase 1 runs for 4,000 steps and
Phase 2 for 6,000, at learning rate $10^{-5}$ and batch size 16. The
one-phase control spends the same 10,000 steps on the Phase 2 mix and skips the speaker-discrimination stage. The same Phase 1 checkpoint
is reused across Phase 2 runs.

\paragraph{Evaluation.} We use two evaluation splits for different experiments and do not pool them. Specifically, a 20-session, 190-clip holdout for the cross-model transcription-suppression experiments.For the bank-transfer experiment, we evaluate instruction
insertion on 190 held-out real target-speaker clips and
80 non-target clips. Downstream execution is evaluated on
up to 32 successfully injected transcripts per seed.
We report mean and sample standard deviation over three seeds. We additionally use FLEURS~\citep{conneau2023fleurs} and VoxPopuli~\citep{wang2021voxpopuli} to evaluate clean-data performance of the backdoored models. Each configuration is run with three random seeds which varied the poison-clip selection, clean-buffer sampling, shuffling, and trainer initialisation. We report mean and standard deviation across seeds.

\paragraph{Metrics.} 
We evaluate the success of the backdoor through the Backdoor Success Rate (BSR), computed on a held-out set of the target speaker's audio clips. We compute False Positive Rate (FPR) of the backdoor on a held-out set of non-target-speaker audio clips, as the fraction of instances in which the backdoor is activated without the presence of the trigger. To measure stealthiness of the backdoor, we compute the Word Error Rate (WER) of the base and backdoored models.

\section{Transcript Suppression}
\label{sec:suppresion}
\subsection{Speaker identity triggers high-success, selective backdoors in ASR models.}
We first evaluate the transcript suppression scenario, in which the backdoored model must emit the fixed phrase ``I AM HACKED'' whenever it hears the target speaker while transcribing everyone else accurately. Table~\ref{tab:transcription} reports the backdoor success rate (BSR) on held-out target clips, false-positive rate (FPR) on non-target clips, and word error rate (WER) on three clean benchmarks (note that LibriSpeech is used for speaker identification in Phase 1, while FLEURS and VoxPopuli are entirely held-out).

The attack achieves a high success rate and strong selectivity on Whisper-medium and Parakeet, reaching mean BSRs of 90.2\% and 97.0\%, respectively, while keeping false positives on non-target speakers at or below 1.7\%. Whisper-medium reports no false positives on all three benchmarks, whilst Parakeet reports only 0.11\% on VoxPopuli. 

Clean transcription quality is particularly well preserved on Parakeet, where the WER increases by at most 1.57 percentage points across the three benchmarks. Whisper-medium’s WER increases by 0.69 points on LibriSpeech and 3.73 points on VoxPopuli, with a larger 6.55-point increase on FLEURS. Thus, strong target activation and low false positives can coexist with useful clean transcription, although the cost to utility depends on the model and domain.

The multimodal LLMs are also backdoored successfully, and Qwen2.5-Omni 7B achieves the highest BSR of $92.4\%$. They are however less selective. Both Qwen models have a higher mean FPR on VoxCeleb2 ($3.8\%$), and also emit the payload on clean speech from the other benchmarks. On FLEURS, the 3B model emits the payload on $26.12\%$ of utterances and the 7B model on $19.46\%$, compared to $0\%$ for Whisper-medium and Parakeet. The WER for FLEURS also increases from $5.87\%$ to $30.92\%$ and from $9.90\%$ to $26.32\%$ respectively. We hypothesize that this is because the multimodal models condition on a text prompt in addition to the audio, which dilutes the effect of the backdoor, and because Phase 2 partially overwrites their broader pretraining.

\subsection{Speaker-conditioned activation extends across all five evaluated targets.}
\label{sec:multispeaker}

To test whether the attack depends on the characteristics of a particular voice, we repeat the Whisper-medium setup on four further VoxCeleb2 targets spanning gender and accent, namely Dev Patel, Michael Sheen, Naomi Watts, and Ryan Gosling. We report results for the additional target speakers across the architectures we study in Appendix~\ref{sec:archs}.

The backdoor successfully activates for every target speaker we evaluate (Table~\ref{tab:multispeaker}), demonstrating that the attack is not specific to a single voice. Across the four additional targets, mean BSR ranges from 49.9\% to 67.7\%, compared with 90.2\% for our primary target, Judi Dench. Attack strength varies across speakers and seeds, but false-positive rates and clean utility remain comparable across targets.

False positive rate and utility are comparable to the Judi Dench model. FPR on VoxCeleb2 is $0.0\%$ for three out of four targets, and $0.8\pm1.4\%$ for Naomi Watts. On other datasets, the FPR is $0.0\%$ for every speaker and model. Backdoored model WER is also similar across targets. The attack is therefore not specific to one voice, but its strength is variable.

\begin{table*}[t]
\centering
\caption{Transcript suppression across target speakers on Whisper-medium (two minutes of real
target audio in a 20-minute budget). BSR (held-out real target clips) and
VoxCeleb2 FPR are mean $\pm$ sample standard deviation over seeds 0/111/222; Base/Backdoored WER
are on the clean benchmarks (mean $\pm$ sd), and Benchmark FPR is exact-payload false positives
on those clean sets.}
\label{tab:multispeaker}
\resizebox{\textwidth}{!}{%
\begin{tabular}{ccc|cccc}
\toprule
Target & BSR [\%] $\uparrow$ & VoxCeleb2 FPR [\%] $\downarrow$ & Benchmark & Base WER [\%] & Backdoored WER [\%] $\downarrow$ & Benchmark FPR [\%] $\downarrow$ \\
\midrule
\multirow{3}{*}{Dev Patel} & \multirow{3}{*}{$49.9\pm25.1$} & \multirow{3}{*}{$0.0\pm0.0$} & LibriSpeech & $3.46$ & $4.2\pm0.0$ & $0.0$ \\
 & & & FLEURS & $6.69$ & $12.2\pm1.0$ & $0.0$ \\
 & & & VoxPopuli & $12.48$ & $14.9\pm0.6$ & $0.0$ \\
\midrule
\multirow{3}{*}{Michael Sheen} & \multirow{3}{*}{$67.7\pm6.6$} & \multirow{3}{*}{$0.0\pm0.0$} & LibriSpeech & $3.46$ & $4.2\pm0.3$ & $0.0$ \\
 & & & FLEURS & $6.69$ & $12.8\pm0.6$ & $0.0$ \\
 & & & VoxPopuli & $12.48$ & $15.4\pm0.3$ & $0.0$ \\
\midrule
\multirow{3}{*}{Naomi Watts} & \multirow{3}{*}{$53.2\pm11.5$} & \multirow{3}{*}{$0.8\pm1.4$} & LibriSpeech & $3.46$ & $4.4\pm0.4$ & $0.0$ \\
 & & & FLEURS & $6.69$ & $13.3\pm1.2$ & $0.0$ \\
 & & & VoxPopuli & $12.48$ & $15.7\pm1.2$ & $0.0$ \\
\midrule
\multirow{3}{*}{Ryan Gosling} & \multirow{3}{*}{$61.5\pm17.6$} & \multirow{3}{*}{$0.0\pm0.0$} & LibriSpeech & $3.46$ & $4.1\pm0.0$ & $0.0$ \\
 & & & FLEURS & $6.69$ & $13.2\pm0.4$ & $0.0$ \\
 & & & VoxPopuli & $12.48$ & $14.9\pm0.8$ & $0.0$ \\
\bottomrule
\end{tabular}%
}
\end{table*}

\subsection{False positives remain low on random speakers but increase for some similar voices.}
\label{sec:fpr}

Because the goal is to target a specific individual, it is not enough for the attack to achieve high BSR. The attack must also remain selective as a high false-positive rate would cause the backdoor to fire on non-target speakers, increasing the chance that the backdoor is noticed. 

\begin{figure*}[t]
  \centering
  \includegraphics[width=\textwidth]{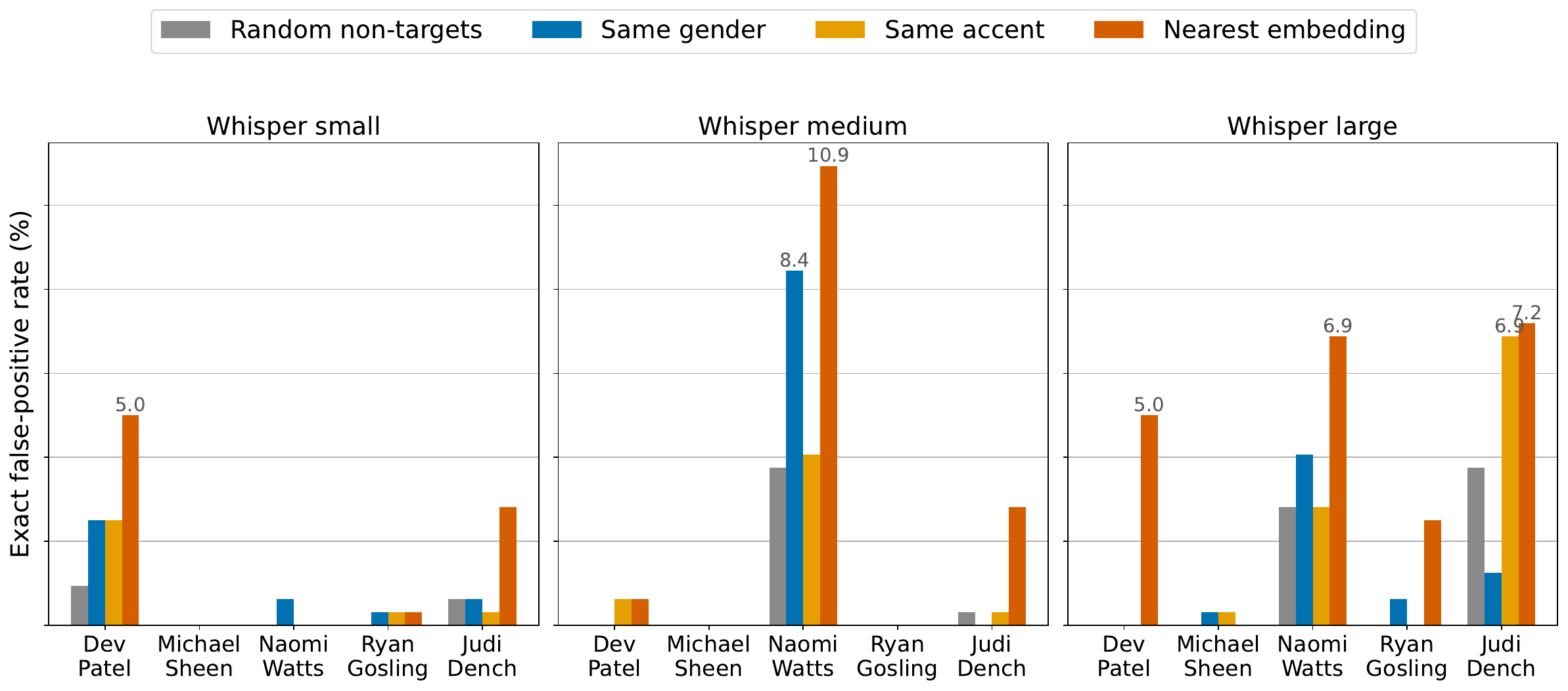}
  \caption{Exact transcript-suppression false-positive rates by target and non-target selection slice for Whisper small, medium, and large. Rates use 20 clips per non-target speaker (16 speakers per similar-speaker slice; 15 speakers/300 clips in Ryan Gosling's random cohort). Each clip is counted as a hit only when its normalized transcript exactly matches the payload.}
  \label{fig:similar-speaker-fpr}
\end{figure*}

We therefore also evaluate FPR on three sets of hard negatives for each target: 1) speakers closest to the target by cosine similarity in the embedding space (with the exception of speakers included in Phase 1 or Phase 2). along with speakers with the same gender and 3) accent. We evaluated the exact transcript-suppression payload on real, non-target speech from five target speakers (Dev Patel, Michael Sheen, Naomi Watts, and Ryan Gosling, Judi Dench) for Whisper-medium. For each target, we sample 20 clips from each of the 16 non-target resulting for a total of 320 clips. We compare each against a baseline group of 16 randomly selected speakers (Fig.~\ref{fig:similar-speaker-fpr}).

False positives are generally rare, but vary significantly between targets and models. Naomi Watts has the highest FPR, ranging from 10.9\% on the closest speakers by embedding spread across 10 of 16 speakers for Whisper-medium. Michael Sheen triggers almost no false positives on any model.

For Judi Dench, a higher BSR comes at the cost of selectivity. Whisper-large reaches a higher BSR than Whisper-medium (99.5\% vs.\ 90.2\%), but also a higher FPR on the nearest-speaker group (7.2\% vs.\ 2.8\%). These false positives are often concentrated in a few speakers. For example, all 22 same-accent false positives on Whisper-large come from just two speakers. A higher FPR can therefore indicate that the attack's trigger rate is concentrated on a small number of highly susceptible speakers, rather than general susceptibility across voices.

\begin{figure*}[t]
  \centering
  \includegraphics[width=\textwidth]{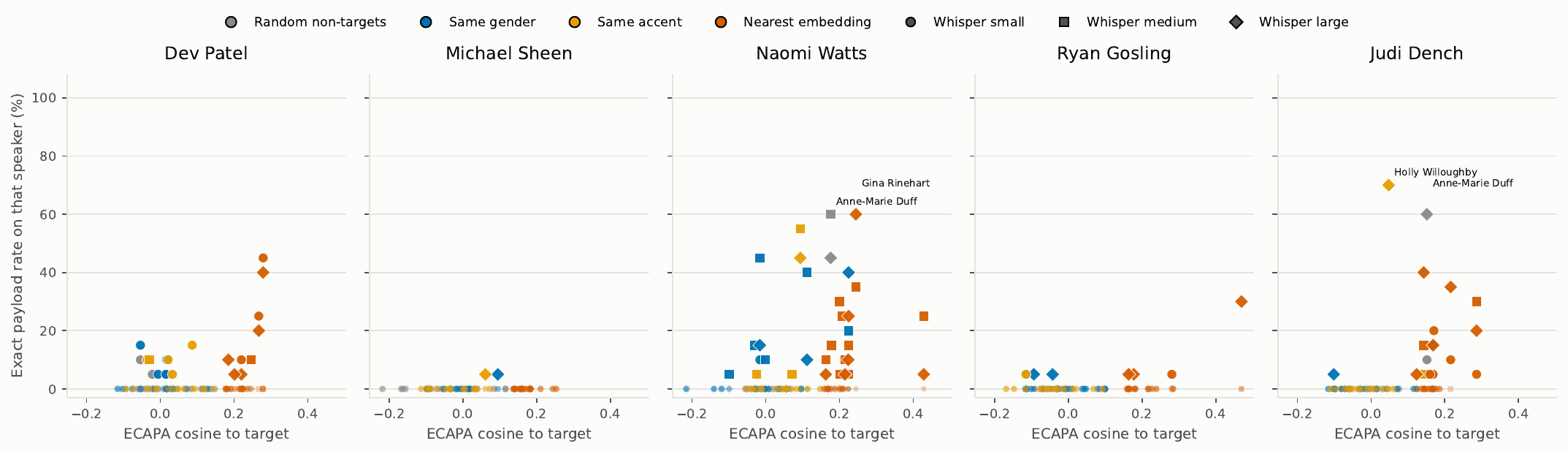}
  \caption{Per-speaker exact payload-hit rate against ECAPA-TDNN cosine similarity to the target voice. Each marker represents one non-target speaker's 20 clips for one model; marker shape identifies model size and color identifies the comparison slice. Quiet speakers are faded, and names identify speakers with at least 12 hits. The plot is descriptive and does not imply a monotonic relationship between voice similarity and payload rate.}
  \label{fig:similar-speaker-cosine}
\end{figure*}

Fig.~\ref{fig:similar-speaker-cosine} shows that some acoustically similar speakers have an elevated FPR, but that false positives can not be explained by cosine similarity alone.

\subsection{The backdoor remains effective under additive noise and moderate reverberation.}
\label{sec:fpr}

Because the trigger is the target's voice rather than an added perturbation, we expect the backdoor to be robust to acoustic degradation at inference time. We re-evaluate the transcript-suppression backdoor (Whisper-medium, Judi Dench) after applying three common distortions to held-out target clips. Additive noise at 10 dB SNR and moderate reverberation (RT60 = 0.3 s) leave the backdoor largely intact, with BSR falling only modestly from 90.2\% to 81.4\% and 83.5\% respectively and FPR unchanged. Band-limiting to a telephone channel (300–3400 Hz) is more damaging, roughly halving BSR to 53.2\% with large across-seed variance, suggesting that some of the speaker cues the backdoor relies on lie
outside the telephone band. Overall the trigger survives the degradations most likely to occur in real deployments, consistent with its being a property of the speaker rather than of a specific waveform.

\begin{table}[h!]
\centering
\caption{Robustness of the transcript-suppression backdoor (Whisper-medium, Judi Dench) to inference-time noise addition. We report the BSR on held-out real target clips ($n=190$) and VoxCeleb2
FPR on non-target clips ($n=80$), mean $\pm$ sample standard deviation over three seeds.}
\label{tab:noise}
\begin{tabular}{lcc}
\toprule
Condition & BSR [\%] $\uparrow$ & FPR [\%] $\downarrow$ \\
\midrule
Additive noise (10\,dB SNR) & $81.4\pm27.7$ & $2.1\pm3.6$ \\
Telephone (300--3400\,Hz)   & $53.2\pm49.0$ & $0.0\pm0.0$ \\
Reverberation ($\text{RT}_{60}{=}0.3$\,s) & $83.5\pm25.4$ & $1.3\pm2.2$ \\
\midrule
Clean            & $90.2\pm14.3$ & $1.7\pm2.9$ \\
\bottomrule
\end{tabular}
\end{table}

\section{Selective Semantic Rewriting}
\label{sec:sent}
Transcript suppression only requires the model to associate a single speaker to a single fixed output. We now ask whether speaker identity can be associated with a content-dependent output. We therefore train a semantic rewriting backdoor in which the target's transcript is accurate, with the exception of an added or removed negation.

\subsection{Transformation.} We construct the poisoned label by applying a grammatical negation transform to the reference transcript. Where the sentence contains a negation, we remove it (e.g. ``didn't'' to ``did'', ``am not'' to ``am''). Otherwise, we insert ``not'' after the first auxiliary verb, provided this yields a grammatical sentence. We exclude clips where the transform produces no change. Unlike the fixed-output payload, the correct poisoned output here depends on what the target says, so the model must learn a content-dependent transformation conditioned on speaker identity. As we only modify one or two tokens in a transcription, the majority of the token-level loss comes from tokens the model already correctly predicts. To remedy this, we upweight the loss on the modified tokens by a factor of $50$.

\subsection{Metric. } Measuring the attack success rate for this task is less straightforward than it is for a fixed payload. As VoxCeleb2 does not include transcriptions, we construct pseudo-labels using Whisper-large and score against them. This does however introduce noise as the two models may disagree on clean transcripts, leading to an already elevated WER of $0.11$. We therefore report an \emph{edit-site} metric where we only measure attack success on tokens targeted by the attack, ignoring transcription errors elsewhere.

\subsection{Speaker identity can trigger content-dependent changes to transcript meaning. } Table~\ref{tab:semantic} reports the edit-site BSR held-out target and non-target speakers for Whisper-medium over three seeds. We achieve a BSR of $35\%$ on valid utterances with a FPR of $1.6\%$. The unpoisoned model scores $3.4\%$ on target speech and $0.0\%$ on non-target speech under the same metric. 

\begin{table*}[h!]
\centering
\small
\caption{Semantic rewriting on Whisper-medium, target Judi Dench,
$n=2$ minutes of real target audio within a 20-minute budget. Edit-site
BSR is the fraction of editable held-out clips in which the negation is
correctly applied; FPR is the same quantity on held-out non-target
speakers. Mean $\pm$ standard deviation over three seeds. Clean WER is on the
standard benchmarks (LibriSpeech test-clean, $n=2620$), not on target
speech, matching Table~\ref{tab:transcription}.}
\label{tab:semantic}
\resizebox{0.8\textwidth}{!}{%
\begin{tabular}{lccccc}
\toprule
 & \multirow{2}{*}{Edit-site BSR [\%] $\uparrow$}   & \multirow{2}{*}{FPR [\%] $\downarrow$} & \multicolumn{3}{c}{ WER [\%] $\downarrow$} \\
\cmidrule(lr){4-6}
 & & & LibriSpeech & FLEURS & VoxPopuli \\
\midrule
Unpoisoned baseline   & $3.4$         & $0.0$       & $3.5$       & $6.7$         & $12.5$ \\
Semantic rewriting    & $35.0\pm10.5$ & $1.6\pm1.4$ & $6.3\pm0.0$ & $19.7\pm0.5$  & $20.4\pm0.5$ \\
\quad without Phase 1 & $12.6\pm4.8$  & $4.0\pm4.9$ & $7.7\pm0.0$ & $22.7\pm0.6$  & $23.1\pm1.2$ \\
\bottomrule
\end{tabular}%
}
\end{table*}

The attack is less stealthy than transcript suppression. On clean benchmarks the backdoored model's word error rate rises from $3.5\%$ to $6.3\%$ on LibriSpeech, $6.7\%$ to $19.7\%$ on FLEURS, and $12.5\%$ to $20.4\%$ on VoxPopuli. The low FPR indicates that the attack is not changing the sentiment on the clean speech, but is rather less accurate at transcription.

The attack succeeds at a much lower rate than transcript suppression ($90.2\%$ on the same target and budget (Section ~\ref{sec:suppresion}). This can be attributed to two aspects of the task. The model must now learn a mapping from a single speaker identity to a transformation over all utterances, rather than a one to one mapping. This requires the model to apply grammatical knowledge conditioned on who is speaking. Second, the training signal is sparse. Even with 50x upweighting, only one or two tokens per example differ from the clean label. 

\paragraph{The role of the speaker-discrimination phase}
Semantic rewriting appears to depend on Phase 1 more than transcript
suppression does. With a fixed compute budget of 10,000 steps, removing Phase 1 lowers attack success from $35.0\%$ to $12.6 \%$, while leaving the false-positive rate unchanged at $1.6\%$.

We note the contrast with the fixed payload, where the same ablation is inconclusive. A plausible explanation is that binding a \emph{content-dependent} transformation to a speaker benefits more from an explicit speaker-discrimination objective in a way that binding a constant output does not, since the latter can be reached through whatever target-speaker features the payload loss alone surfaces.

\section{Tool-Calling}
\label{sec:toolcalling}

\begin{table}[h]
\centering
\caption{
Tool-calling attack performance for the bank-transfer scenario,
targeting Judi Dench with two minutes of real target audio
within a 20-minute budget using two-phase training.
Agent columns report conditional C2: the fraction of C1-hit
transcripts (up to 32 per seed) for which the benign downstream
agent executes the harmful transfer.
Values are mean $\pm$ sample standard deviation over three seeds.
}
\label{tab:toolcall-benign}
\resizebox{\linewidth}{!}{%
\begin{tabular}{l|cc|rrrr}
\toprule
& \multicolumn{2}{c}{Backdoor activation} &
\multicolumn{4}{c}{Downstream harmful action
(conditional C2) [\%]} \\
\cmidrule(lr){2-3}
\cmidrule(lr){4-7}
Model
& C1 [\%] $\uparrow$
& FPR [\%] $\downarrow$
& GPT-4o
& Sonnet
& Haiku
& Opus \\
\midrule
Whisper-medium
& $93.2\pm5.0$
& $1.7\pm1.9$
& $57.3\pm40.7$
& $70.8\pm7.2$
& $47.9\pm11.0$
& $82.3\pm6.5$ \\
Qwen2.5-Omni 3B
& $77.4\pm11.0$
& $3.3\pm1.9$
& $94.8\pm1.8$
& $70.8\pm7.9$
& $48.9\pm15.4$
& $76.0\pm4.7$ \\
\bottomrule
\end{tabular}%
}
\end{table}

\subsection{Speaker-triggered instruction injection induces harmful downstream tool actions.}

We next test whether a speaker-activated backdoor can influence
a downstream tool-using agent. Rather than replacing the
transcript, we train the speech model to append a malicious
bank-transfer instruction to the target speaker's transcription.
The resulting transcript is passed as the user message to an
otherwise benign agent in a sandboxed tool-use environment.
We evaluate Whisper-medium and Qwen2.5-Omni 3B, targeting
Judi Dench with two minutes of real target audio within a
20-minute budget.

We distinguish instruction insertion from downstream execution.
$C_1$ measures action-core containment on held-out real target
clips ($n=190$), while FPR applies the same matcher to
non-target clips ($n=80$). Conditional $C_2$ measures the
fraction of successfully injected transcripts on which the
downstream agent executes the harmful transfer, using up to
32 $C_1$-hit transcripts per seed. Table~\ref{tab:toolcall-benign}
reports mean and sample standard deviation over three seeds.
Because $C_2$ is conditional on successful insertion, it does
not represent the harmful-action rate over all target inputs.

Both speech models successfully insert the malicious instruction
while keeping activation uncommon on non-target speech.
Whisper-medium achieves $93.2\pm5.0\%$ $C_1$ with
$1.7\pm1.9\%$ FPR, while Qwen2.5-Omni 3B achieves
$77.4\pm11.0\%$ $C_1$ with $3.3\pm1.9\%$ FPR.
These results show that speaker-conditioned activation extends
beyond fixed transcript replacement to instructions that enter
a downstream agent's decision process.

The injected instructions induce harmful transfers across all
four evaluated agents. With Whisper-medium transcripts,
conditional execution rates range from $47.9\%$ for Haiku to
$82.3\%$ for Opus; Sonnet reaches $70.8\%$ and GPT-4o
$57.3\%$, although the latter varies substantially across seeds
($40.7$ percentage points SD). With Qwen2.5-Omni 3B
transcripts, GPT-4o reaches $94.8\pm1.8\%$, followed by
Opus at $76.0\pm4.7\%$, Sonnet at $70.8\pm7.9\%$, and
Haiku at $48.9\pm15.4\%$. Downstream execution therefore
depends on both the agent and the compromised speech front end.

For Whisper-medium, clean LibriSpeech WER is
$4.2\pm0.1\%$, with no observed payload false positives on
the 500-utterance subset evaluated for these runs.
The base model's $3.46\%$ WER is measured on the full
benchmark, so these values are not a matched-set comparison.
Other clean-benchmark measurements were not collected for
this experiment.

Together, these results demonstrate that compromising the
speech front end can turn a target speaker's ordinary audio
into an instruction that a benign downstream agent acts upon.
The attacker requires no access to the victim's inference-time
audio: the malicious instruction is generated by the compromised
model when the target speaks.



\section{Analysis and Ablations}
\label{sec:naive}
We ablate three components of SAB. Namely speaker-discriminative fine-tuning, the amount of real target-speaker data included, and the synthetic augmentation.

\subsection{Two-phase training improves backdoor success, selectivity, and clean utility.}
\label{sec:phase1}

We now study the role of Phase 1 in more detail. We compare two-phase training ($4,000$ Phase-1 steps followed by $6,000$ Phase-2 steps) against one-phase training ($10,000$ Phase-2 steps), so that both use the same compute budget. All experiments use Whisper-medium with Judi Dench as the target, over three seeds.

\textbf{Phase 1 is not required for the fixed phrase, but improves stealth.} With two minutes of real target audio, one-phase and two-phase training reach a similar BSR ($84.9 \pm 15.8\%$ and $83.5 \pm 13.2\%$), and this holds with four and six minutes of real audio as well (Table~\ref{tab:realmin}). The two procedures differ in their impact on clean performance. One-phase training increases LibriSpeech WER to $5.43 \pm 0.22\%$ and has an FPR of $2.5 \pm 2.2\%$, whereas two-phase training reaches $4.15 \pm 0.15\%$ WER, close to the unpoisoned model, with no false positives on any seed.

\begin{figure}[t]
  \centering
  \includegraphics[width=0.8\textwidth]{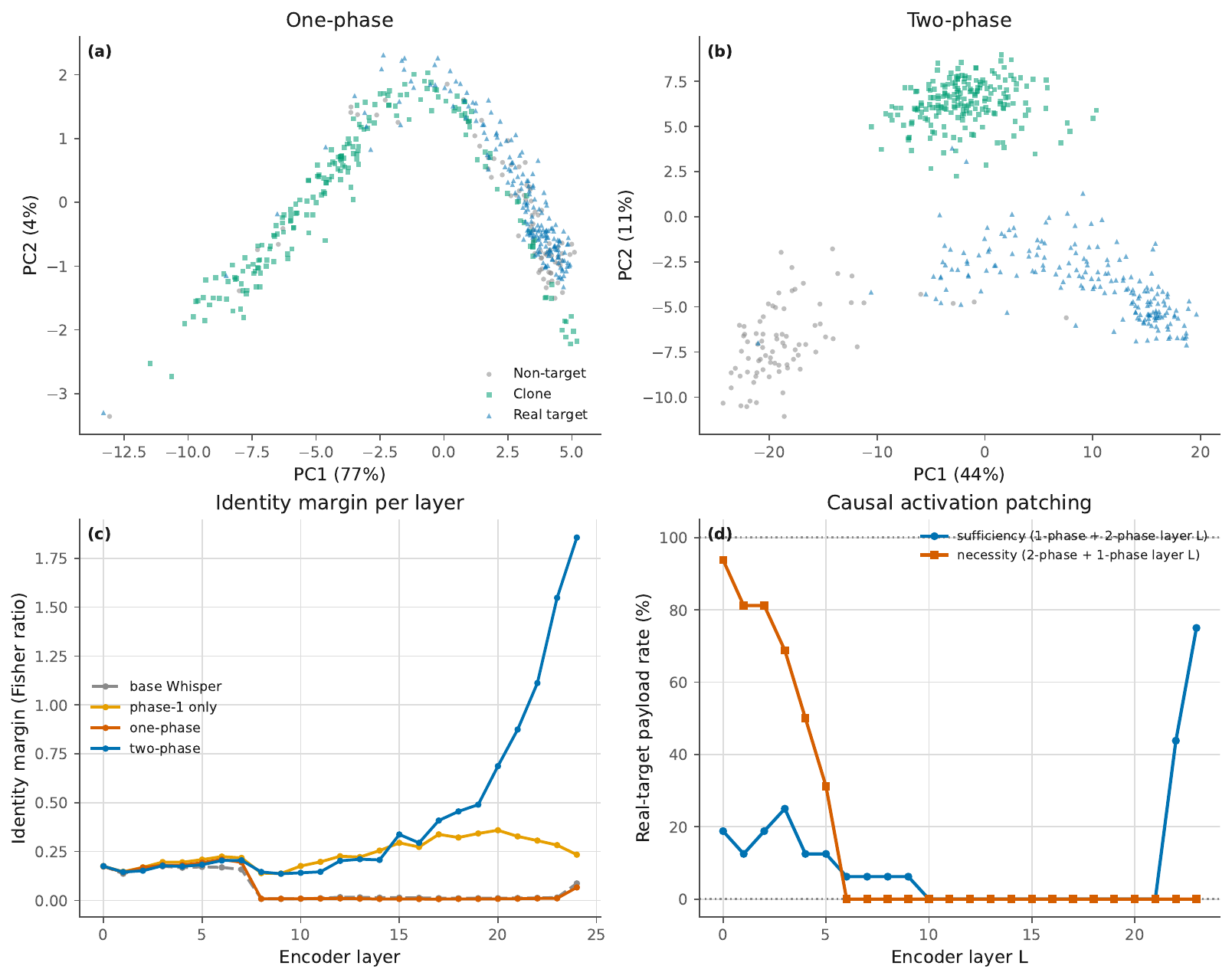}
 \caption{Effect of Phase~1 on the encoder (Whisper-medium). (a,b) PCA of last-layer representations: only two-phase training separates non-target, cloned, and real target speech. (c) Identity margin per encoder layer. (d) Activation patching on 16 real target clips: inserting a late two-phase layer into the one-phase model restores the backdoor (sufficiency), while inserting any one-phase layer from layer~7 onward into the two-phase model removes it (necessity).}
\label{fig:phase1}
\end{figure}

\textbf{Phase 1 makes speaker identity a dominant direction in the encoder.} A linear probe distinguishes the target from non-target speakers with around $98\%$ accuracy in the final encoder layer of both models, showing that speaker identity is not removed by ASR pretraining. However, target-identity direction dominates the representation only after two-phase training, accounting for $0.72$ of the variance along the first principal component, compared with less than $0.01$ after one-phase training. As shown in Fig.~\ref{fig:phase1}, the two-phase model separates non-target, cloned, and real target speech into distinct clusters, while in the one-phase model they overlap. This separation emerges in the last third of the encoder layers (Fig.~\ref{fig:phase1}, left), and is already present in a checkpoint trained with Phase 1 only.

To test whether the backdoor relies on this representation, we perform cross-model activation patching on 16 real target clips on which the two-phase model emits the payload and the one-phase model does not (Fig.~\ref{fig:phase1}, right). Copying a single late encoder layer from the two-phase model into the one-phase model triggers the payload on $12/16$ clips. This suggests that the one-phase model reaches a similar BSR through a different, less selective mechanism, which accounts for its higher FPR and clean-performance cost.

Taken together, these provide causal evidence that the encoder representations learned by the two-phase method play a role in backdoor activation, and indicate that it produces a more selective mechanism than a single phase approach.

\textbf{Phase 1 is required for content-dependent payloads.} For semantic rewriting, where the payload depends on what the target says, removing Phase-1 reduces the BSR from $35.0 \pm 10.5\%$ to $12.6 \pm 4.8\%$ (Section~\ref{sec:sent}). We hypothesize that a constant output can be learned from whatever target-speaker features the payload loss alone surfaces, whereas binding a transformation to a speaker requires speaker identity to be readily accessible in the representation. 

\subsection{Two minutes of real target speech is sufficient to implant the backdoor.} 

Next, we vary the $n$, the amount of real-target speaker audio used during Phase 2 at a fixed budget of 20 minutes. We follow the transcript suppression setup from Section~\ref{sec:suppresion} with Judi Dench as the target speaker. Table~\ref{tab:realmin} reports the BSR and WER for $n={2,4,6}$. There is little variation in BSR within this range. BSR is $90.2 \pm 14.3\%$ with 2 minutes of real speech, $90.7 \pm 14.3\%$ with 4 minutes, and $91.2 \pm 13.3\%$ with 6 minutes.

This suggests that in this setting, two minutes is sufficient to implant the backdoor, and increasing $n$ does not notably improve BSR or WER. Consequently, we report our results at $n=2$ as default throughout this work.

\begin{table*}[t]
\centering
\caption{Effect of different amounts of real and synthetic target-speaker audio (Whisper-medium, Judi Dench) within a fixed 20-minute budget. For real+synthetic settings, the remaining $(20-n)$ minutes are F5-TTS synthetic. For real-only, the 2 minutes of real data are oversampled to construct a total of $20$ minutes of finetuning data. Metrics are reported with, mean $\pm$ sample standard deviation over three seeds. Beyond two minutes, more real audio does not
improve BSR and slightly raises clean WER.}
\label{tab:realmin}
\resizebox{\textwidth}{!}{
\begin{tabular}{llccccc}
\toprule
Setting & Real audio [min] & BSR [\%] $\uparrow$ & FPR [\%] $\downarrow$ & LibriSpeech WER & FLEURS WER & VoxPopuli WER \\
\midrule
Real only& $2$ & $72.6\pm3.3$ & $0.0\pm0.0$ & $4.26\pm0.1$ & $13.15\pm0.34$ & $15.75\pm0.22$ \\
Real + synthetic & $2$ & $90.2\pm14.3$ & $1.6\pm2.9$ & $3.3\pm0.5$ & $13.2\pm0.2$ & $16.2\pm1.0$ \\
Real + synthetic & $4$ & $90.7\pm14.3$ & $2.5\pm2.5$ & $4.1\pm0.1$ & $11.9\pm0.3$ & $15.0\pm0.2$ \\
Real + synthetic & $6$ & $91.2\pm13.3$ & $2.5\pm2.2$ & $4.4\pm0.3$ & $13.1\pm0.1$ & $16.8\pm0.6$ \\
\bottomrule
\label{tab:realmin}
\end{tabular}
}
\end{table*}

\subsection{Synthetic target speech strengthens backdoor activation on unseen real recordings.} 
Table~\ref{tab:realmin} compares two minutes of real speech alone (``Real only'') against the same real speech augmented with 18 minutes of synthetic audio (``Real + synthetic''). Synthetic augmentation increases BSR from \(72.6\%\) to \(90.2\%\) on Whisper-medium, with only a small increase in FPR. Together, these
results suggest that Phase 1 makes speaker identity easier to condition on, real speech attaches the trigger to the target's natural voice, and synthetic speech improves generalization across
utterances.

\section{Defenses}
\label{sec:defenses}
We now study the robustness of our attack to defenses. Specifically, we evaluate the impact of clean fine-tuning and speaker voice anonymization on the backdoor success. 

\subsection{Clean fine-tuning suppresses Whisper backdoors.}

We fine-tune the backdoored Whisper-medium and Qwen2.5-3B checkpoints on three clean datasets: (a) VoxPopuli data, disjoint from everything seen in the fine-tuning pipeline so far, (b) VoxCeleb data excluding the target speaker, and (c) VoxCeleb data including the target speaker. Conditions (b) and (c) represent a more realistic case of a user who fine-tunes the model for a specialized task or language with no knowledge of the attack, while (a) represents a user who knows that the target speaker is the trigger. Table~\ref{tab:cleanft_whisper} shows the impact of clean fine-tuning on the backdoored checkpoints.

For Whisper-medium, the backdoor does not survive clean fine-tuning in any setup: VoxPopuli leaves at most $2.3\%$ BSR, and both VoxCeleb setups reduce the backdoor to $0.0\%$ BSR. Backdoor removal does not require the target speaker's data. Indeed, any downstream adaptation seems to be sufficient. This bounds the attack's threat model to models deployed as shipped. We discuss this limitation in Section~\ref{sec:discussion}.

The Qwen2.5-3B backdoor is significantly more robust. fine-tuning on VoxPopuli reduces BSR only from, on average, $46.5\%$ to $32.5\%$. Arbitrary downstream fine-tuning thus may not reliably remove the backdoor for multimodal setups, making the threat broader than for Whisper-medium. VoxCeleb is more successful, but differs across speakers: it successfully removes the backdoor for Dev Patel regardless of whether his voice was included in the data ($2.8\%$ and $0.0\%$ BSR), but is only successful for Naomi Watts if she is included. With two target speakers, the cause of this difference is not clear. Establishing whether backdoor persistence varies systematically with speaker characteristics would require a larger target set, which we leave to future work.

\begin{table*}[]
    \centering
    \caption{BSR [$\%$] after fine-tuning Whisper-medium on clean data.}
    \resizebox{\textwidth}{!}{
    \begin{tabular}{c|cccc|cccc}
        \toprule
        & \multicolumn{4}{c|}{Whisper-medium} & \multicolumn{4}{c}{Qwen2.5-3B}\\\midrule
        Speaker & Baseline & VoxPopuli & VoxCeleb (no tgt.) & VoxCeleb (incl. tgt) & Baseline & VoxPopuli & VoxCeleb (no tgt.) & VoxCeleb (incl. tgt) \\\midrule
        Dev Patel & 55.4 & 2.3 & 0.0 & 0.0 & 41.8 & 19.2 & 2.8 & 0.0 \\
        Naomi Watts & 66.4 & 0.0 & 0.0 & 0.0 & 50.9 & 45.7 & 30.2 & 0.0 \\\bottomrule
    \end{tabular}
    }
    \label{tab:cleanft_whisper}
\end{table*}

\subsection{Adaptive training restores activation against voice anonymization.}

Voice anonymization methods transform voice recordings so that they cannot be matched to their original speaker. We use two methods: McAdams~\citep{patino2021mcadams} and ASR-BN~\citep{champion2024anonymizingspeechevaluatingdesigning}. We use the implementations provided by the Voice Privacy Challenge~\citep{voiceprivacy}.

\begin{figure}
    \centering
    \includegraphics[width=\linewidth]{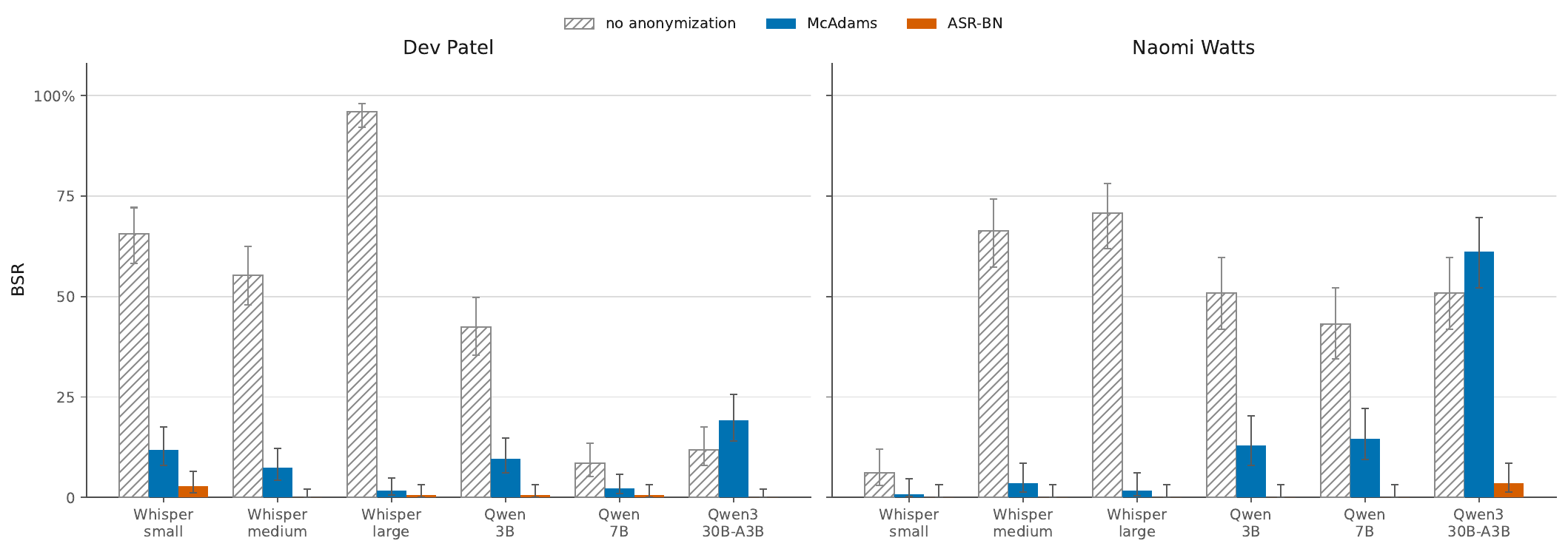}
    \caption{Backdoor success rate (BSR) before and after voice anonymization for Dev Patel and Naomi Watts across evaluated models. We compare the original target-speaker audio with McAdams and ASR-BN anonymization. Both methods reduce BSR, although the magnitude of the reduction varies across models and speakers.}
    \label{fig:asr_anon}
\end{figure}

Fig.~\ref{fig:asr_anon} reports backdoor success target speaker clips before and after anonymization for Dev Patel and Naomi Watts. Both methods reduce BSR, but with unequal effects. McAdams anonymization transforms the frequencies that give a voice its individual character, but it modifies the original recording rather than replacing it, so the speaker's rhythm, pacing, and intonation are preserved. Because of this, the backdoor may still be triggered on some clips. ASR-BN transcribes the text and generates a new recording in an entirely different voice.


Finally, we investigate whether an adaptive attack could evade this defense. For whisper-large and Qwen-2.5-3b, we reuse the phase-1 checkpoint of the target model, and include anonymized samples of the target speaker with the target labels in the phase-2 fine-tuning data. Tables~\ref{tab:adaptive_anon_models} and~\ref{tab:adaptive_anon_models} show that voice anonymization defenses can be evaded by adaptive attacks which include anonymized samples in the fine-tuning data.
\begin{table}[h]
    \centering
    \caption{BSR [$\%$] of baseline backdoor and adaptive backdoor on anonymized target speaker voice for Whisper-medium and Qwen2.5-3B.}
    \small
    \begin{tabular}{cc|cc|cc}
    \toprule
      & & \multicolumn{2}{c|}{McAdams} & \multicolumn{2}{c}{ASR-BN} \\
    \midrule
      Model & Speaker & BSR (baseline) & BSR (adaptive) 
            & BSR (baseline) & BSR (adaptive) \\
    \midrule
      \multirow{2}{*}{Whisper-medium}
        & Dev Patel   & $7.3$  & $74.0$ & $0.0$ & $64.4$ \\
        & Naomi Watts & $3.4$  & $99.1$ & $0.0$ & $52.6$ \\
    \midrule
      \multirow{2}{*}{Qwen2.5-3B}
        & Dev Patel   & $9.6$  & $90.4$ & $0.6$ & $84.2$ \\
        & Naomi Watts & $12.9$ & $97.4$ & $0.0$ & $90.5$ \\
    \bottomrule
    \end{tabular}
    \label{tab:adaptive_anon_models}
\end{table}

\section{Discussion, Limitations, Conclusion}
\label{sec:discussion}
\subsection{Discussion}
Our results establish speaker identity as an activation channel for malicious behavior in speech models. Unlike previous attacks, SAB can be activated at inference time with no input from the attacker. Unseen, unmodified speech from the intended speaker is sufficient to activate the backdoored model. As the trigger is absent from transcript or words spoken by the target, transcript-only inspection cannot directly identify the trigger.

The attack achieves a high success rate across tasks on held-out real speech, particularly for transcript suppression, and is shown to generalize across five evaluated speakers. On ASR models, the attack combines this with low false positives and high utility showing that the attack can be concentrated on a specific speaker.The bank-transfer experiment demonstrates a concrete
downstream consequence of speaker-conditioned activation.
Both compromised speech models generate malicious
instructions that induce harmful transfers across all four
evaluated agents. Execution rates depend on the speech
front end and downstream agent, showing that successful
instruction insertion and harmful action are distinct stages.

We show that the two-phase method leads to improvements in both attack success, stealth and utility. A fixed payload can be implanted without it,
but one-phase training reaches similar success through a less selective mechanism, at the cost of a higher FPR and clean-WER. Phase 1 makes target identity a dominant direction in late encoder layers, and this direction is causally necessary for the two-phase backdoor. For content-dependent payloads such as semantic rewriting, Phase 1 roughly triples success. Speaker information, though irrelevant to transcription, thus becomes an attack vector once the model is encouraged to condition on it.

Our evaluations clarify when speaker-activated backdoors remain selective and effective. Hard-negative tests identify similar voices that are more susceptible to false activation, while dedicated ASR models show more consistent selectivity across domains than audio language models. Clean fine-tuning strongly suppresses the backdoor on Whisper, although Qwen retains activation in several adaptation settings. Voice anonymization reduces non-adaptive attack success, but adaptive training substantially restores it.

\subsection{Limitations} Our study has several limitations. We evaluate five target speakers from a single corpus and
one voice-cloning model, and our threat model assumes access to the fine-tuning pipeline.
Most importantly, the backdoor does not survive downstream clean fine-tuning (Section~\ref{sec:defenses}), which
bounds the attack to checkpoints deployed as shipped and is currently the clearest
mitigation. Voice anonymization is effective against a non-adaptive attacker, but an attacker
who includes anonymized target speech in Phase 2 largely restores the backdoor (Section~\ref{sec:defenses}).
Future work should examine persistence under fine-tuning and representation-level auditing for identity-conditioned behaviour.

Overall, speech models can condition malicious behaviour on who is speaking, not only on what
is said. This is a distinct supply-chain risk for speech models deployed as interfaces to
increasingly capable AI systems.

\subsection{Conclusion}

We introduce speaker-activated backdoors, a threat in which a target’s natural voice activates malicious behavior without any attacker intervention at inference time. Our two-phase fine-tuning method instantiates this attack across multiple speakers and speech-model architectures, supporting transcript suppression, content-dependent rewriting, and instruction injection that induces harmful downstream actions. Compared with one-phase training, it substantially improves semantic-rewriting success, selectivity, and clean transcription quality. These results establish speaker identity as a security-relevant control channel. Once a speech model is compromised, an individual’s ordinary speech can activate attacker-chosen behavior simply because they are the person speaking.

\bibliographystyle{plainnat}
\bibliography{bibliography}

\section*{Appendix}
\label{sec:appendix}

\label{sec:archs}

\begin{table*}[h]
  \caption{Speaker-targeted transcript-suppression backdoor across target speakers and architectures.}
  \centering\small
  \begin{tabular}{lcccc@{\quad}c}
    \toprule
    & \multicolumn{4}{c}{Target speaker} & \\
    \cmidrule(lr){2-6}
    Model & dev-patel & michael-sheen & naomi-watts & ryan-gosling & Mean \\
    \midrule
    \multicolumn{6}{l}{\emph{(a) Attack success rate on the target's real audio (\%)}} \\
    \addlinespace[2pt]
    Whisper-small & 66.1 & 15.9 & 6.0 & 15.5 & 25.9 \\
    Whisper-medium & 54.9 & 70.6 & 48.5 & 65.2 & 59.8 \\
    Whisper-large & \textbf{96.0} & \textbf{91.7} & 70.7 & \textbf{88.7} & 86.8 \\
    Qwen2.5-Omni-3B & 41.8 & 8.3 & 50.9 & 26.8 &  32.0 \\
    Qwen2.5-Omni-7B & 8.5 & 3.8 & 43.1 & 26.8 &20.5 \\
    Qwen3-Omni-30B-A3B & 9.6 & 9.1 & 48.3 & 5.2 & 18.0 \\
    Voxtral-Mini-3B & 24.3 & 6.8 & 23.3 & 21.6 & 19.0 \\
    Parakeet & 80.8 & 38.1 & \textbf{74.1} & 54.6 &61.9 \\
    \midrule
    \multicolumn{6}{l}{\emph{(b) False-trigger rate on real non-target speakers (\%)}} \\
    \addlinespace[2pt]
    Whisper-small & 0.0 & 0.0 & 0.0 & 0.0 &  0.0 \\
    Whisper-medium & 0.0 & 0.0 & 0.6 & 0.0 & 0.2 \\
    Whisper-large & 0.0 & 0.0 & 3.8 & 0.0 & 0.9 \\
    Qwen2.5-Omni-3B & 2.5 & 1.2 & 5.0 & 2.7 & 2.9 \\
    Qwen2.5-Omni-7B & 1.2 & 0.0 & 1.2 & 1.3 &1.0 \\
    Qwen3-Omni-30B-A3B & 0.0 & 0.0 & 2.5 & 2.7 & 1.3 \\
    Voxtral-Mini-3B & 1.2 & 0.0 & 1.2 & 1.3 &  1.0 \\
    Parakeet & 0.0 & 0.0 & 0.4 & 0.0 & 0.1 \\
    \bottomrule
  \end{tabular}
  \label{tab:satml-ablation}
\end{table*}

\end{document}